\documentclass[10pt,letterpaper]{article}

\usepackage{cogsci}

\cogscifinalcopy 
\usepackage[natbibapa]{apacite}

\usepackage{amssymb}
\usepackage{amsmath}
\usepackage{siunitx}

\usepackage[hidelinks]{hyperref}

\usepackage{graphicx}
\usepackage{pslatex}
\usepackage{apacite}
\usepackage{float} 

\usepackage{todonotes}
\usepackage{tikz}
\DeclareRobustCommand*\numcircledtikz[1]{%
  \protect\tikz[baseline=(char.base)]{%
    \protect\node[shape=circle,draw,inner sep=1.2pt] (char) {#1};%
  }%
}
\DeclareMathOperator*{\argmax}{\arg\max}

\title{From Curiosity to Competence: How World Models Interact with the Dynamics of Exploration}

\author{\textbf{Fryderyk Mantiuk$^{1,*}$}(fryderyk.mantiuk@student.uni-tuebingen.de), \textbf{Hanqi Zhou$^{1, 2,*}$}(hanqi.zhou@uni-tuebingen.de) \\\& \textbf{Charley M. Wu$^{1,2}$} \\
$^1$ Human and Machine Cognition Lab, University of T\"ubingen, T\"ubingen, Germany\\
$^2$ Department of Computational Neuroscience, Max Planck Institute for Biological Cybernetics\\
$^*$ shared authorship
}

\begin{document}

\maketitle

\begin{abstract} 
What drives an agent to explore the world while also maintaining control over the environment? 
From a child at play to scientists in the lab, intelligent agents must balance curiosity (the drive to seek knowledge) with competence (the drive to master and control the environment). 
Bridging cognitive theories of intrinsic motivation with reinforcement learning, we ask how evolving internal representations mediate the trade-off between curiosity (novelty or information gain) and competence (empowerment). 
We compare two model-based agents using handcrafted state abstractions (Tabular) or learning an internal world model (Dreamer). 
The Tabular agent shows curiosity and competence guide exploration in distinct patterns, while prioritizing both improves exploration. 
The Dreamer agent reveals a two-way interaction between exploration and representation learning, mirroring  the developmental co-evolution of curiosity and competence. 
Our findings formalize adaptive exploration as a balance between pursuing the unknown and the controllable, offering insights for cognitive theories and efficient reinforcement learning.

\end{abstract}

\section{Introduction}
Picture a child playing with two brand-new toys. One lights up predictably when a button is pressed, while the other flickers randomly, indifferent to the child’s actions. Which toy captivates the child longer? The answer depends not just on novelty or control but on a deeper interplay between two fundamental drives: \emph{curiosity}, which compels us to seek new knowledge, and \emph{competence}, which drives us to leverage what we know to influence our environment \citep{meltzoff2012learning, nussenbaum2019reinforcement}. 

This highlights a broader puzzle about human exploration: we are drawn to both the \textit{unknown}, in our quest to reduce uncertainty, and the \textit{predictable}, in our desire to exert influence over the world \citep{driveforknowledge, nelson2005finding, ten2024curious}. Curiosity-related drives, such as \emph{novelty} and \emph{information gain}, push us to reduce uncertainty and build better mental models of the world \citep{ten2024curious, modirshanechi2022taxonomy, gottlieb2013information}. Competence-related drives, such as \emph{empowerment} and skill learning, motivate us to predict and control outcomes \citep{klyubin2005empowerment, salge2014empowerment, gopnik2024empowerment, poli2020infants, aubret2023information}. 

At first glance, these drives might seem sequential: first curiosity for learning, then competence to act effectively. But the reality is far more recursive. Consider a child learning to walk: only by developing competence in mobility can they access new environments to satisfy their curiosity. Conversely, the child’s curiosity about more distant places may fuel their persistence in mastering locomotion. This bidirectional relationship raises a critical question: How do curiosity and competence co-evolve, as agents build and refine their understanding of the world?

\begin{figure*}[ht]
    \centering
    \includegraphics[width=0.8\linewidth]{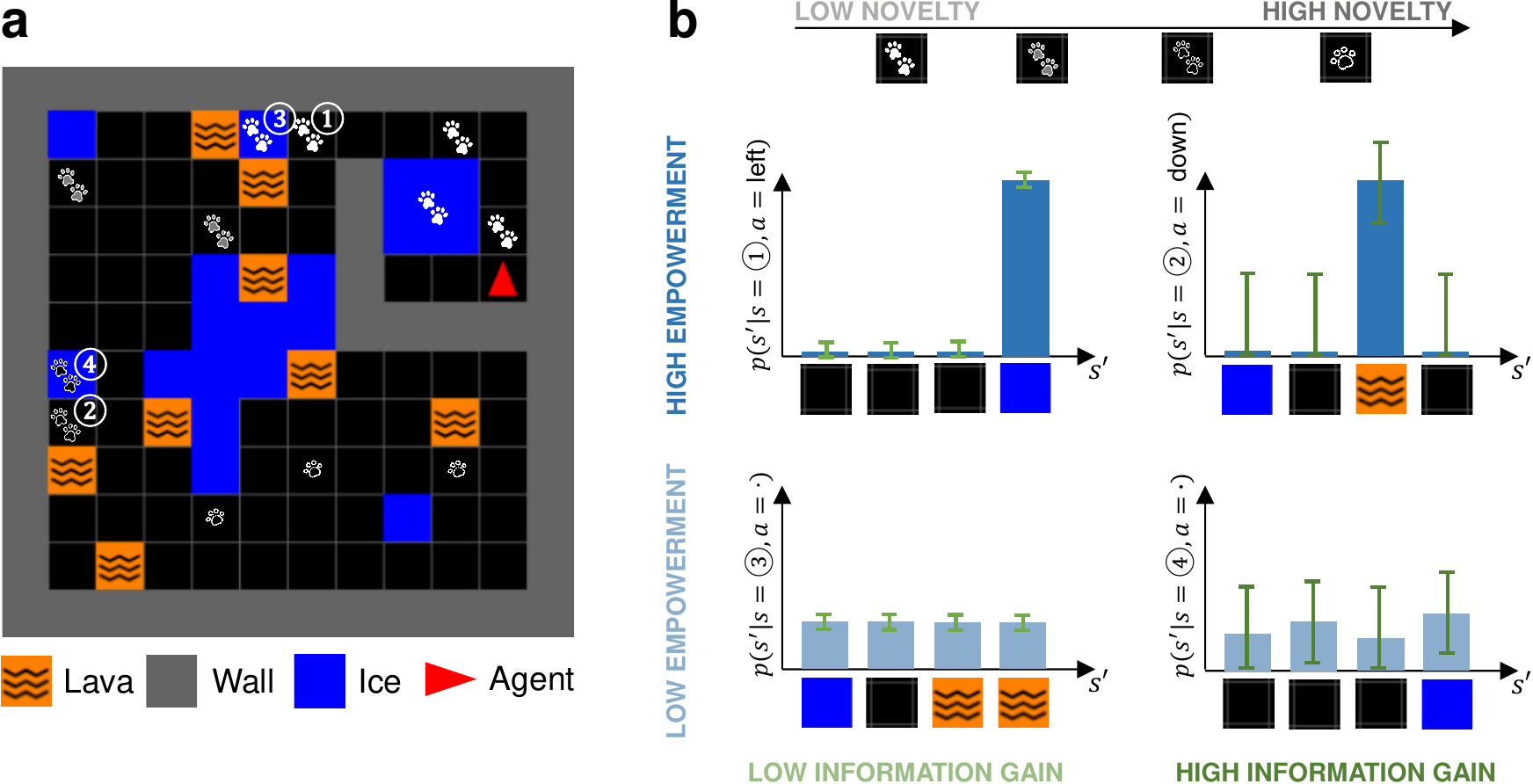}
    \caption{\textbf{Task overview}. \textbf{a}) The mixed-state playground has three different cells: lava, wall, and ice. The agent (red triangle) starts from the upper right corner and has to explore the environment while avoiding death (falling into lava). 
    \textbf{b}) Novelty depends only on the visitation frequency for each state, regardless of the state types. Information gain is low for transitions which the agent has already learned to predict well (cells \numcircledtikz{1} and \numcircledtikz{3}), and high for transitions that the agent is less confident about (cells \numcircledtikz{2} and \numcircledtikz{4}). Empowerment is low in slippery ice cells where the agent cannot control where it will end up (cells \numcircledtikz{3} and \numcircledtikz{4}), and high in states with many predictable action outcomes (cells \numcircledtikz{1} and \numcircledtikz{2}).}
    \label{fig:env_and_info_gain_vs_empowerment}
    \vspace{-1em}
\end{figure*}

Existing reinforcement learning (RL) agents struggle with this balance \citep{ocana2023overview, dulac2021challenges, gruaz2024merits}. Curiosity-driven agents, typically prioritizing novelty or information gain, often fall prey to the ``noisy TV problem'', getting distracted by random, uncontrollable stimuli that offer no meaningful opportunities for mastery \citep{schmidhuber1991curious, burda2018exploration, pathak2017curiosity}. Conversely, competence-focused agents, which maximize empowerment or control, often assume fixed world models, neglecting how exploration might reshape those models \citep{du2023can, brandle2023empowerment, lidayan2025intrinsically}. Humans, in contrast, dynamically adjust their focus: a toddler confined to a crib may inspect every toy in meticulous detail, while the same child, once mobile, might prioritize breadth over depth, racing to explore new rooms. What enables this adaptive prioritization?

A key piece of the puzzle lies in the agent’s ``world model''---the internal representation of environmental dynamics that guides predictions and decisions \citep{ha2018world, hafner2023mastering}. When uncertainty is high, curiosity dominates, driving exploration to refine the model. As confidence grows, competence takes precedence, leveraging the model to achieve goals. Yet this is not a one-way transition. Competence unlocks new frontiers for curiosity (e.g., learning to walk expands the horizon for exploration), while curiosity generates the knowledge needed for higher-order competence (e.g., understanding physics enables tool use). This creates a feedback loop: world models shape exploration strategies, which in turn reshape the models themselves.

\paragraph{Goal and scope.}
In this work, we investigate how agents balance curiosity and competence during learning and exploration. We ask: under what conditions should exploration prioritize curiosity versus competence, and how do these priorities shift as world models evolve? By integrating insights from cognitive science and RL, we aim to shed light on the mechanisms that allow humans and artificial agents to navigate the fine line between learning and doing, uncertainty and mastery, and curiosity and competence.

Specifically, we compare three intrinsic motivations in sparse-reward environments: \textit{novelty} (collecting diverse experiences), \textit{information gain} (epistemic uncertainty reduction), and \textit{empowerment} (perceived control over future states). We simulated and analyzed two different agents on grid worlds: (1) a \textbf{Tabular} Q-learning agent whose state representations are predefined by handcrafted features; and (2) a \textbf{Dreamer} world-model agent \citep{hafner2023mastering} that learns a latent representation of the world during exploration.

With simulations in grid worlds, we analyze how each intrinsic motivation drives divergent exploration patterns, showing that they are neither functionally redundant nor universally optimal. 
Empirically, we show how environmental structure (stochastic dynamics and the presence of fatal danger) and the agent's evolving world model systematically modulate the efficacy of exploration strategies.
Together, the findings clarify how safe and effective exploration could arise from balancing curiosity-driven information seeking with competence-driven control, each offering distinct advantages depending on the environmental context.

\section{Methods} 
In this paper, we investigate the interactions between intrinsic motivations and model-based state representations using RL agents in a grid-based environment \cite[MiniGrid;][]{MinigridMiniworld23}. The following sections describe the key elements---the environment to explore, the agents that explore it,
and intrinsic rewards that determine \textit{how} the agents explore the environment.

\subsection{Environment}
Our environments are based on MiniGrid \citep{MinigridMiniworld23}, a 2D grid-world framework which encourages exploration. 
To mimic real-world challenges, we first extend MiniGrid and design a heterogeneous playground (Fig.~\ref{fig:env_and_info_gain_vs_empowerment}a) which has three state types: lava (imposing irreversible penalties, terminating the episode), ice (stochastic transitions), and barriers (constraining mobility; e.g., walls or moving balls). These states operationalize fundamental trade-offs between risk, uncertainty, and control that are also observed in naturalistic decision-making.

The agent starts in the upper right corner, requiring it to first exit through a bottleneck of walls and lava. If it succeeds, it could later find a larger area to explore in the bottom right, either via a shortcut through an icy and stochastic patch or by taking a longer but more reliable route.

\subsection{Agent-environment interactions}
Interactions between the agent and the environment are defined as a Partially Observable Markov Decision Process (POMDP) \citep{sutton1998reinforcement}. At each time step, the agent receives an observation~$o \in \mathcal{O}$, which may include the agent’s immediate surroundings, such as walls, objects, or other entities within its field of view. 

Based on its internal representation $z$ of the current observation $o$, the agent selects action~$a \sim \pi(a \mid z)$ from a discrete action space given its policy~$\pi$. 
The agent can face four directions and has three basic actions: move forward, or turn left or right.
After executing an action, the agent transitions to a new state~$s'$ according to the environment’s transition dynamics~$p(s' \mid s, a)$. The agent may also receive an extrinsic reward~$r$ for reaching a goal state (see Simulation 2: generalization phase), although all agents initially need to rely on intrinsic rewards alone for the majority of the training, to build up their model of the environment.

\begin{figure}[t]
    \centering
    \includegraphics[width=\linewidth]{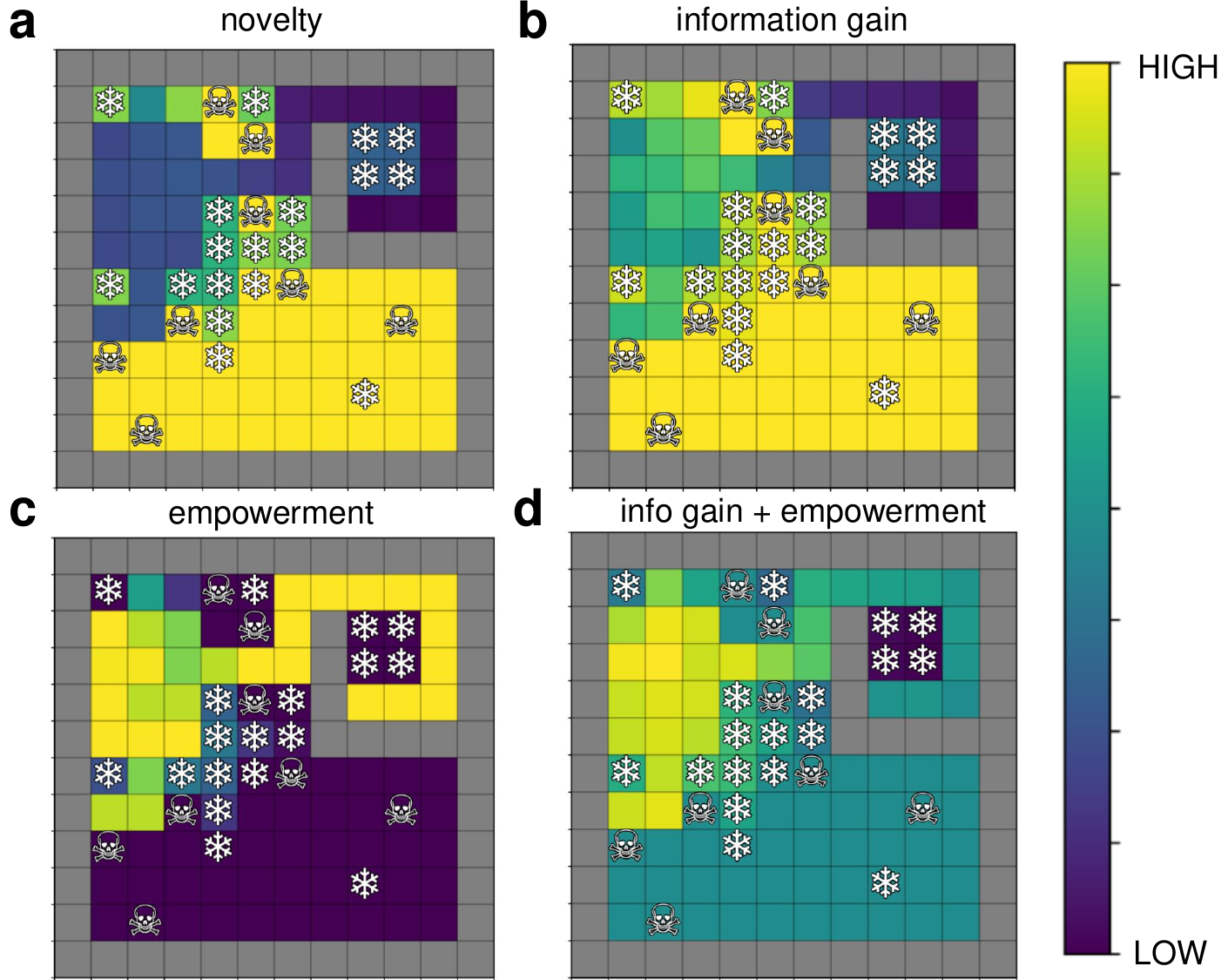}
    \vspace{-1em}
    \caption{\textbf{Intrinsic motivation heatmaps}. A snapshot of each motivation metric in all tiles in Simulation 1 (skull represents fatal lava tiles, snowflake represents stochastic ice tiles). \textbf{a}) Novelty and \textbf{b}) information gain drive the agent to explore, but fail to recognize that the stochasticity of the slippery ice cannot be reduced through experience. \textbf{c}) Empowerment learns to avoid ice and lava, preferring the predictable dynamics of neutral states instead.
    \textbf{d}) Summing information gain and empowerment obtains the highest rewards in areas with controllable yet not fully explored dynamics, sensibly prioritizing reducible over irreducible uncertainty. }
    \label{fig:heatmaps}
    \vspace{-1em}
\end{figure}

\subsection{Agents} 
A core challenge in studying exploration is that real-world agents must simultaneously learn \emph{what parts of the world to focus on} (state representations) and \emph{how the world works} (transition dynamics). Prior work often assumes that intrinsic motivations are computed on fixed and hand-crafted state representations \citep{du2023can, lidayan2025intrinsically}, or by learning these representations and policies (or value functions) separately \citep{burda2018exploration, ferrao2025world}. 
However, this sidesteps a critical question: how do agents explore effectively when their understanding of ``states'' is formed by their experiences? 

To evaluate how intrinsic motivations are impacted by changing internal representations, we conduct two progressive simulations: (1) a Tabular agent with predefined state representations to probe the interplay between different intrinsic motivations under fixed representations, and (2) a Dreamer agent that autonomously learns latent state representations from observations in the pixel space to further understand the role of these evolving representations. 

\paragraph{Tabular agent.}
This first set of simulations provides a controlled baseline for isolating the dynamics of learning a transition model of the environment (i.e., model-based RL) and providing intuitions of the exploratory behavior under different motivations. Here, state representations are predefined, allowing the agent to focus on refining its transition model. 
 
Our simple fixed representation reduces states to (position, orientation) tuples $z=(x, y, \theta) \in \mathbb{N}^3$ where $x$ and $y$ are the agent's grid coordinates and $\theta$ is the facing direction. On top of this, we can estimate a count-based transition model of the environment dynamics:

\begin{equation}
p(z'\mid z,a) = \frac{\alpha N(z,a,z') + 1}{ \sum_{z' \in \mathcal{Z} }N(z,a,z') + 1}
\label{eq:modeltransitions}
\end{equation}
where $N(z,a,z')$ counts transitions from state $z$ to $z'$ under action $a$, scaled by an update factor $\alpha=100|Z|$, which allows the agent to quickly update from the uninformative uniform prior towards the counts it actually experienced. 

Following \citet{gruaz2024merits}, the agent learns Q-values with fast model-based updating through prioritized sweeping with the learned dynamics (Eq.~\ref{eq:modeltransitions}):

\begin{equation}
Q(z,a) = \sum_{z^\prime \in \mathcal{Z}} p(z^\prime\mid z,a) \left[r(z,a,z^\prime) + \gamma \max_{a^\prime} Q(z^\prime,a^\prime) \right],
\label{eq:Qupdates}
\end{equation}
where the discount factor~$\gamma = \sqrt[|\mathcal{Z}|/2]{0.5}$ is defined by the number of states~$|\mathcal{Z}|$, and the immediate reward~$r(z,a,z^\prime)$ is computed as the respective intrinsic objective of novelty, information gain or empowerment.
To select actions, the agent uses a greedy policy over the Q-values, $p(a\mid z) = \argmax\limits_a Q(z,a)$.

\paragraph{Dreamer agent.}
To test how learned representations mediate exploration, we employ DreamerV3 \citep{hafner2023mastering} as a world-model reinforcement learning agent that constructs discrete latent states from raw pixel observations without task-specific supervision. 
Specifically, the world model $f_{\mathrm{wm}}$ has two main components: an encoder $p_\phi(z' \mid z,a, o)$, which encodes raw observations $o'$ into discrete latent states $z$, and a dynamics predictor $p_\phi(z' \mid z, a)$, which predicts $z'$ from previous $z$ and action $a$. These latent states distill sensory inputs into abstract variables (e.g., encoding ``door states'' as locked/unlocked), analogous to unsupervised category formation.
For implementation and simulations, we largely follow the same architecture and hyperparameters as in \citet{ferrao2025world}, but use the partial pixel-based observation provided by MiniGrid, rather than the semantic observations which they used. 

\subsection{Intrinsic motivation} 
Intrinsic motivation drives agents to explore and learn, independent of extrinsic rewards. We focus on three key types: \emph{novelty}, \emph{information gain}, and \emph{empowerment}, commonly used in both machine learning and psychology. 

\noindent \textbf{Novelty} encourages exploration by prioritizing unfamiliar states \citep[Fig.~\ref{fig:heatmaps}a;][]{berlyne1950novelty, schwartenbeck2013exploration, dubey2020reconciling}. We define novelty as the state surprise:
\begin{equation}
   R_\text{Novelty}(z,a,z')=-\log \frac{N(z)}{\sum_{\tilde{z} \in \mathcal{Z}} N\left(\tilde{z}\right)},
    \label{eq:novelty}
\end{equation} 
where $N(z)$ is a count of how often the agent has seen the state representation~$z$. This metric focuses on exploring less-visited states, but is agnostic about state-action dynamics (Fig.~\ref{fig:env_and_info_gain_vs_empowerment}b). Since DreamerV3 has a discrete latent space, we can use the same count-based novelty for both Tabular agents and Dreamer agents.

\noindent \textbf{Information gain} drives agents to perform actions in states where they are uncertain about the outcome, aiming to improve their world model \citep[Fig.~\ref{fig:heatmaps}b;][]{nelson2005finding, gottlieb2013information, still2012information, oudeyer2007intrinsic}. 
This improvement is often quantified by how much the uncertainty in the world model is reduced after observing a transition.
For the tabular world model, we compute the \textit{predicted} information gain, following past work \citep{little2013learning, modirshanechi2022taxonomy, gruaz2024merits, kolossa2015computational}: 

\begin{equation}
    R_{\text{IG}_\text{Tabular}} (z,a,z') = \mathbb{E}_{\hat z' \sim p(Z'\mid z,a)} \left[ \operatorname{KL}\left (p(Z'\mid z,a,\hat z') \mid\mid p(Z' \mid z,a) \right) \right ]
    \label{eq:predicted_information_gain_tabular}
\end{equation} 
 
The Dreamer agent requires a slightly different computation, since computing the expectation over all possible future states would require an intractable amount of forward passes through the dynamics predictor. Here, instead of computing the predicted information gain, we we can compute it \textit{retrospectively} (after observing a transition), by using the KL divergence between posterior (latent state predicted by encoder that has access to the observation) and prior (latent state predicted by dynamics predictor before seeing the observation) as an information gain reward:

\begin{equation}
R_{\text{IG}_\text{Dreamer}} (z,a,z') = \operatorname{KL}(p_\phi(z' \mid z,a,o') \mid \mid p_\phi(z' \mid z, a))
\label{eq:information_gain_dreamer}
\end{equation}

\noindent \textbf{Empowerment} reflects the degree of control an agent has over its environment \citep[Fig.~\ref{fig:heatmaps}c;][]{klyubin2005empowerment,  gopnik2024empowerment, du2023can, lidayan2025intrinsically} by maximizing the mutual information between its actions and resulting states:
\begin{equation}
\mathfrak{E}(s)=\max_{\omega(a\mid s)} I\left(S^{\prime} ; A \mid  s\right).
\label{eq:empowerment}
\end{equation}
Thus, empowerment encourages an agent to visit states where it has the greatest influence over future outcomes, where both a greater diversity of outcomes and less stochasticity in the mapping of actions to outcomes contribute to more empowerment (Fig.~\ref{fig:env_and_info_gain_vs_empowerment}c). Empowerment thus prioritizes states where the agent has the highest agency, independent of uncertainty or visitation frequency. 

Again, we compute this intrinsic motivation over the latent representations $z$ rather than the observations $o$ or true states $s$.
In the Tabular agent, we compute 1-step empowerment using the Blahut-Arimoto algorithm, which iteratively optimizes the action distribution $\omega$ to maximize the mutual information between actions and resulting states \citep{dupuis2004blahut}. The empowerment reward is then simply:

\begin{equation}
R_{\text{Emp}_\text{Tabular}}(z,a,z') = \mathfrak{E}(z) = \max_{\omega(a\mid z)} I\left(Z^{\prime} ; A \mid  z\right)
\label{eq:empowerment_reward_tabular}
\end{equation}

In the Dreamer agent, applying Blahut-Arimoto to find the optimal action distribution is intractable, so we assume a random policy $\omega = \mathcal{U}$ following \citet{seitzer2021causal} and compute the reward

\begin{equation}
R_{\text{Emp}_\text{Dreamer}}(z,a,z') = I(Z^{\prime};A \mid z) =  H(Z^{\prime}  \mid  z) - H(Z^{\prime}  \mid  A, z).
\label{eq:empowerment_reward_dreamer}
\end{equation}

\noindent \textbf{Combinations} of empowerment with any of the two knowledge-seeking measures into a unified reward for the same agent have not been explored to the best of our knowledge. To evaluate whether competence-seeking in the form of empowerment could benefit from knowledge-seeking (and vice versa), we choose to combine information gain and empowerment in two simple ways, via a sum or a product. This combination is chosen for two reasons: Firstly, from a theoretical perspective, information gain aims to specifically improve the dynamics model---which is necessary to compute empowerment---while novelty only rewards seeing infrequently visited states. Secondly, from an empirical perspective, we observed that information gain resulted in a more thorough exploration than novelty in both Tabular and Dreamer experiments.

\section{Results} 
\subsection{Simulation 1: tabular agent}
\label{sec:qlearning-result}

\begin{figure}[t]
    \centering
    \includegraphics[width=\linewidth]{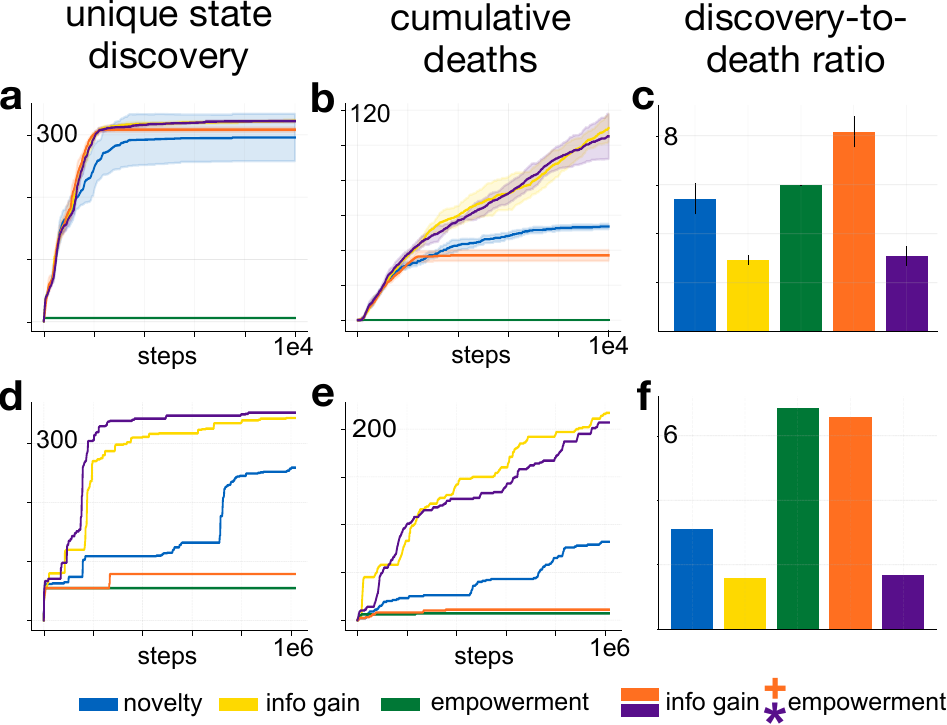}
    \vspace{-1em}
    \caption{\textbf{Exploration patterns} of the Tabular (a-c) and Dreamer agents (d-f) on the mixed-state playground  (Fig.~\ref{fig:env_and_info_gain_vs_empowerment}). \textbf{a, d)} Total number of new states discovered over time. \textbf{b, e)} Total number of deaths over time. \textbf{c, f)} Ratio of discovered states to deaths after $1e4$, and $1e6$ steps respectively.}
    \label{fig:tabular_results}
    \vspace{-1em}
\end{figure}

Each agent was evaluated over 10,000 steps and 5 random rounds using different intrinsic rewards: novelty (Eq.~\ref{eq:novelty}), information gain (Eq.~\ref{eq:predicted_information_gain_tabular} and Eq.~\ref{eq:information_gain_dreamer}), empowerment (Eq.~\ref{eq:empowerment_reward_tabular} and Eq.~\ref{eq:empowerment_reward_dreamer}), along with two combinations of information gain and empowerment (sum and product) to explore their synergy. This allows us to test whether agents can distinguish reducible (epistemic) uncertainty from irreducible (aleatoric) uncertainty in environmental dynamics, by prioritizing useful unknowns (controllable floor cells) over useless unpredictability (e.g., stochastic ice cells).

The results (Fig.~\ref{fig:tabular_results}a-c) show that the most thorough discovery is achieved by a pure information-gain-seeking agent (Fig.~\ref{fig:tabular_results}a). However, this exploration comes at a cost (Fig.~\ref{fig:tabular_results}b): The agent often risks dying in its quest to perfectly predict the stochastic dynamics in ice cells, making it slip into lava. In contrast, empowerment fails to explore and stays in the starting state, avoiding death entirely. 

Crucially, combining information gain and empowerment rewards via a naive sum results in a more balanced approach, leading to a higher discovery-to-death ratio (Fig.~\ref{fig:tabular_results}c): The jointly curiosity- and competence-seeking agent explores most of the environment while avoiding uncontrollability as soon as it recognizes it.

\subsection{Simulation 2: dreamer agent}

We now turn to simulation where the state representations are learned during exploration by a Dreamer world-model agent. Thus the intrinsic motivations are derived from dynamically evolving latent states, allowing for a bi-directional interaction. 
The Dreamer agent can test whether it bypasses perceptual distractions (e.g., a ``noisy TV'' generating irreducible randomness) to instead target controllable uncertainties, such as states where actions reliably alter transitions. 

\begin{figure}[t]
    \centering
    \includegraphics[width=\linewidth]{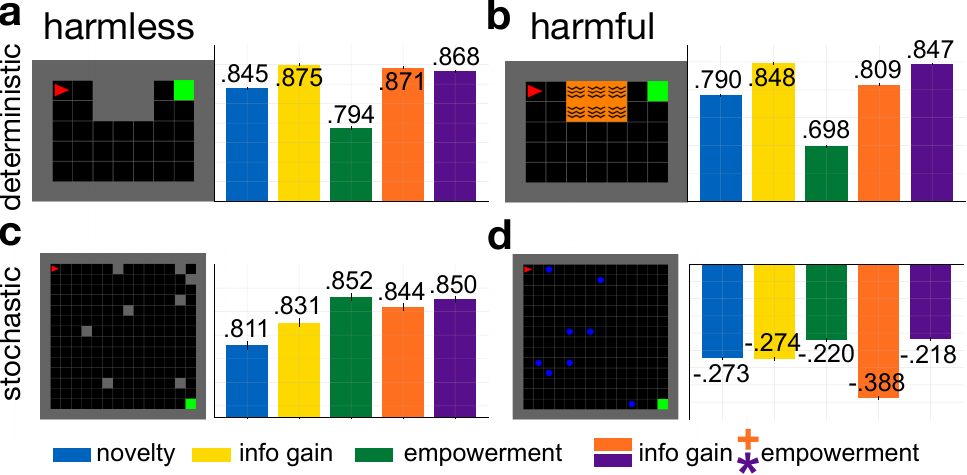}
    \vspace{-1em}
    \caption{\textbf{Generalization performance} of the Dreamer agent on unary-state grid worlds. The unary-state environments are split into $2 \times 2$ categories, together with the agent's generalization performance under different intrinsic motivations. }
    \label{fig:dreamer-generalization}
     \vspace{-1em}
\end{figure}

\begin{figure}[t]
    \centering
    \includegraphics[width=\linewidth]{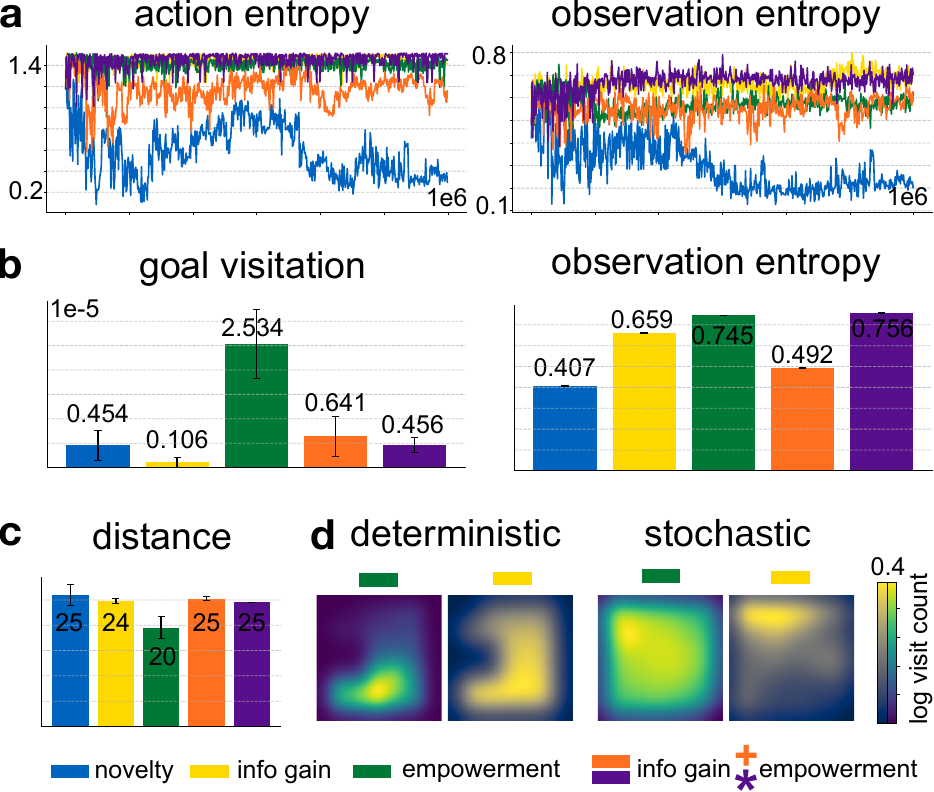}
    \vspace{-1em}
    \caption{\textbf{Exploration patterns} of the Dreamer agent during pretraining. \textbf{a)} Observation and action entropy over time averaged over all environments. \textbf{b)} Extrinsic reward and observation entropy in the stochastic + harmless environment (averaged over time steps), which is the hardest for information gain. \textbf{c)} Distance between the agent and the lower border in deterministic environments. \textbf{d)} Heatmaps of state visitations for empowerment and information gain agents. } 
    \label{fig:dreamer-pattern}
    \vspace{-1em}
\end{figure}

\paragraph{On the mixed-state playground. }
We begin by presenting results comparable to those of tabular agents, evaluating effective exploration (in terms of discovery and deaths) across different motivations in Figure~\ref{fig:tabular_results}d-f. Much like the Tabular agent, we can observe that an agent driven by empowerment alone exhibits poor exploration behavior (Fig.~\ref{fig:tabular_results}d). However, it is able to explore all the states in the corner room where it starts, in contrast to the Tabular agent. Yet, the Dreamer agent with empowerment is still unable to traverse the bottleneck and explore the rest of the environment, contrary to those driven by other motivations. The synergistic effects of the combinations of empowerment and information gain begin to mirror the Tabular setting, but are less pronounced.

\paragraph{On the unary-state grid. }
To better understand the learned world representation and its role in exploration behavior, we further isolate the impact of individual state types. 
We decompose the environment used in Simulation 1 into four variants with orthogonal features (Fig.~\ref{fig:dreamer-generalization}): harmful vs. harmless and deterministic vs. stochastic. The first dimension modulates the safety of the environment, where harmful environments introduce states that result in reward penalties (lava and balls). The second dimension modulates the predictability of the environment: the stochastic environments introduce stochastic transitions in the placement of walls or path-blocking blue balls. This design allows us to test whether \emph{novelty} distinguishes true semantic novelty from continually high combinatorial novelty from random configurations, whether \emph{information gain} overfixates on inherently random (and unlearnable) transitions, and whether \emph{empowerment} can maintain a locus of control when this requires more than simply standing still. 

Our analysis has two phases. (1) Pretraining: The agent learns in four distinct environments using only intrinsic rewards. (2) Generalization: The agent is tested in novel environments (similar properties as pretraining but differing in layout or dynamics) with only extrinsic rewards. 

\noindent\textbf{Generalization performance} is shown in Figure~\ref{fig:dreamer-generalization} (higher is better). 
Performance is highly context-specific: novelty and information gain excel in deterministic environments, whereas empowerment proves more effective in stochastic settings. 
However, hybrid approaches (combining information gain and empowerment through a sum or product) are generally equal to or better than each individual intrinsic motivation, suggesting their synergy may offer a promising compromise. 

\noindent \textbf{Exploration patterns} show that
\emph{Novelty} had the lowest entropy of observations and actions during exploration (Fig.~\ref{fig:dreamer-pattern}a), meaning it encountered the fewest new states and had the lowest diversity of actions. This is because the agent is satisfied with a trivial form of novelty: the world model naturally develops changes in a few dimensions of its learned state representations every time, thus increasing the subjective sense of novelty without translating into concrete exploration.

\emph{Information gain} improved exploration in deterministic contexts, but had trouble in stochastic ones (Fig.~\ref{fig:dreamer-generalization}b). This aligns with our results in the tabular setting: contrary to what one might expect from theory, information gain often conflates (reducible) epistemic uncertainty with (irreducible) aleatoric uncertainty in practice. Focusing on stochastic + harmless environments (Fig.~\ref{fig:dreamer-pattern}b), the information gain agent achieves minimal goal visits (during pretraining) despite having high observation entropy. This is due to the agent fixating on inherently unpredictable transitions (randomly moving walls) rather than on exploration.

\emph{Empowerment} diverges fundamentally from novelty and information gain by prioritizing control over future states.
In deterministic environments, this is maladaptive, with the agents getting stuck in a stable ``comfort zone'' (regions maximally distant from walls or lava or near the starting point). Quantitative analysis of wall-distance distributions further supports this, showing agents systematically position themselves to retain access to multiple future paths (Fig.~\ref{fig:dreamer-pattern}c). Heatmaps of exploration patterns (Fig.~\ref{fig:dreamer-pattern}d) also confirm this tendency, revealing a strong preference for obstacle-sparse regions in deterministic settings (i.e., bottom right).
However, in stochastic environments, an emphasis on controllability becomes adaptive. Here, the empowerment agent abandons its comfort zone to instead roam continuously to maintain influence over outcomes. This contrasts starkly with information gain, which fixates on inherently irreducible aleatoric noise (e.g., tracking random object movements) and becomes trapped near the starting area (upper left corner).

\section{Discussion}
We investigated how intrinsic motivation mechanisms---specifically curiosity (novelty and information gain) and competence (empowerment)---guide exploration in model-based reinforcement learning (RL) agents. By comparing two architectures---a tabular agent with fixed state representations and a Dreamer agent with learned ones---we demonstrate that curiosity and competence play complementary roles in exploration and enhance generalization.

Our results show distinct trade-offs among intrinsic motivation strategies. \textit{Novelty}-driven exploration could get stuck in local optima, going back and forth between a limited set of states. \textit{Information gain} avoided this because it sought out reducible uncertainty which required exploring the entire environment, but it was slowed down in stochastic contexts by mistakenly fixating on irreducible uncertainty. \textit{Empowerment} preferred deterministic dynamics---sometimes hindering exploration by staying in a controllable ``comfort zone'', but sometimes instead aiding exploration by actively avoiding harmful or risky stochasticity to preserve agency, thus avoiding the myopic distractions that limited curiosity-driven strategies. Hybrid strategies combining information gain and empowerment leverage this natural complementarity to achieve better exploration-safety balances in the tabular agent. This synergy was also present in the Dreamer agent with more robust generalization in novel environments, although the effects were less pronounced.

While our simplified grid-worlds enabled precise manipulation of state transitions, they exclude challenges posed by open-ended worlds. Additionally, although our computational agents were designed to isolate core principles of exploration, validating these mechanisms against human behavior \cite[e.g., in video games like Crafter;][]{du2023can, lidayan2025intrinsically} and over developmental timescales \citep{giron2023developmental} remains essential to assess their psychological plausibility. Future work could extend our framework by (1) integrating adaptive motivation strategies that dynamically adapt the weighting of curiosity and competence based on environment type, (2) testing whether our observed empowerment-like ``safety first'' models of behavior could be used to describe human subjects under threat-of-shock paradigms, and (3) scaling these principles to real-world robotics tasks requiring robustness to environmental stochasticity.

In sum, we studied the co-evolution of world model learning and exploration strategies defined by different intrinsic motivation mechanisms. 
Our results revealed that although each intrinsic motivation had context-specific advantages, hybrid strategies yielded synergistic benefits, illustrating the complementarity of curiosity and competence.

 \clearpage 
\section{Acknowledgments}
We thank Alison Gopnik, Aly Lidayan, Eliza Kosoy, Alireza Modirshanechi, Georg Martius and Cansu Sancaktar for helpful discussions. 
The authors thank the International Max Planck Research School for Intelligent Systems (IMPRS-IS) for supporting HZ. 
This research is supported as part of the LEAD Graduate School \& Research Network, which is funded by the Ministry of Science, Research and the Arts of the state of Baden-Württemberg within the framework of the sustainability funding for the projects of the Excellence Initiative II.
This work is supported by the German Federal Ministry of Education and Research (BMBF): Tübingen AI Center, FKZ: 01IS18039A, funded by the Deutsche Forschungsgemeinschaft (DFG, German Research Foundation) under Germany’s Excellence Strategy–EXC2064/1–390727645, and funded by the DFG under Germany's Excellence Strategy – EXC 2117 – 422037984.

\bibliographystyle{apacite}

\setlength{\bibleftmargin}{.125in}
\setlength{\bibindent}{-\bibleftmargin}

\bibliography{CogSci_Template}

\end{document}